\documentclass{article}

\PassOptionsToPackage{table}{xcolor}

\usepackage{lmodern}
\usepackage[T1]{fontenc}
\usepackage[numbers,sort]{natbib}
\usepackage{arxiv}

\usepackage{authblk}
\usepackage{csquotes}
\usepackage{url}
\usepackage{tabularx}
\usepackage{enumitem}
\usepackage{nicefrac}
\usepackage{microtype}
\usepackage{threeparttable}
\usepackage{makecell}
\usepackage{subcaption}
\usepackage{afterpage}

\usepackage{amsmath,amsfonts,bm}

\def\eqref#1{equation~\ref{#1}}

\def\1{\bm{1}}

\DeclareMathAlphabet{\mathsfit}{\encodingdefault}{\sfdefault}{m}{sl}
\SetMathAlphabet{\mathsfit}{bold}{\encodingdefault}{\sfdefault}{bx}{n}

\makeatletter
\def\@copyrightspace{\relax}
\makeatother

\title{OckBench: Measuring the Efficiency of LLM Reasoning}

\author[1]{Zheng Du$^{*}$}
\author[1]{Hao Kang$^{*}$}
\author[2,3]{Song Han}
\author[1]{Tushar Krishna}
\author[3]{Ligeng Zhu}

\affil[1]{Georgia Institute of Technology}
\affil[2]{Massachusetts Institute of Technology}
\affil[3]{NVIDIA}

\date{}

\begin{document}

\maketitle

\begingroup
  \renewcommand\thefootnote{}
  \footnotetext{$^{*}$Equal contribution. Correspond to \href{mailto:zdu@gatech.edu}{zdu@gatech.edu}.}
\endgroup

\begin{abstract}
  Large language models (LLMs), such as GPT-5.5 and Claude Opus 4.7, have pushed the frontier of automated reasoning and code generation. Yet current benchmarks mainly emphasize accuracy and output quality, while overlooking a key dimension: token efficiency. We introduce \textbf{OckBench}, a benchmark designed to jointly evaluate accuracy and token efficiency across reasoning and coding tasks. Our evaluation highlights three takeaways. First, models with similar accuracy can show large variance in full-output tokens, including reasoning and final answers: frontier and open-source models can differ by over \textbf{26$\times$} in average output length. Second, cheaper token cost does not always imply cheaper task cost; verbose smaller models can pay an \textit{Overthinking Tax} that makes them more expensive to serve. Third, we observe \textit{Per-Token Intelligence}: smarter and larger models generally produce shorter responses, reaching correct answers with denser reasoning. These findings provide a concrete roadmap for improving reasoning ability and token efficiency. Ultimately, we argue for a shift in evaluation: tokens should not be multiplied beyond necessity.
\end{abstract}

\afterpage{%
  \begin{figure}[t]
    \centering
    \includegraphics[width=\linewidth]{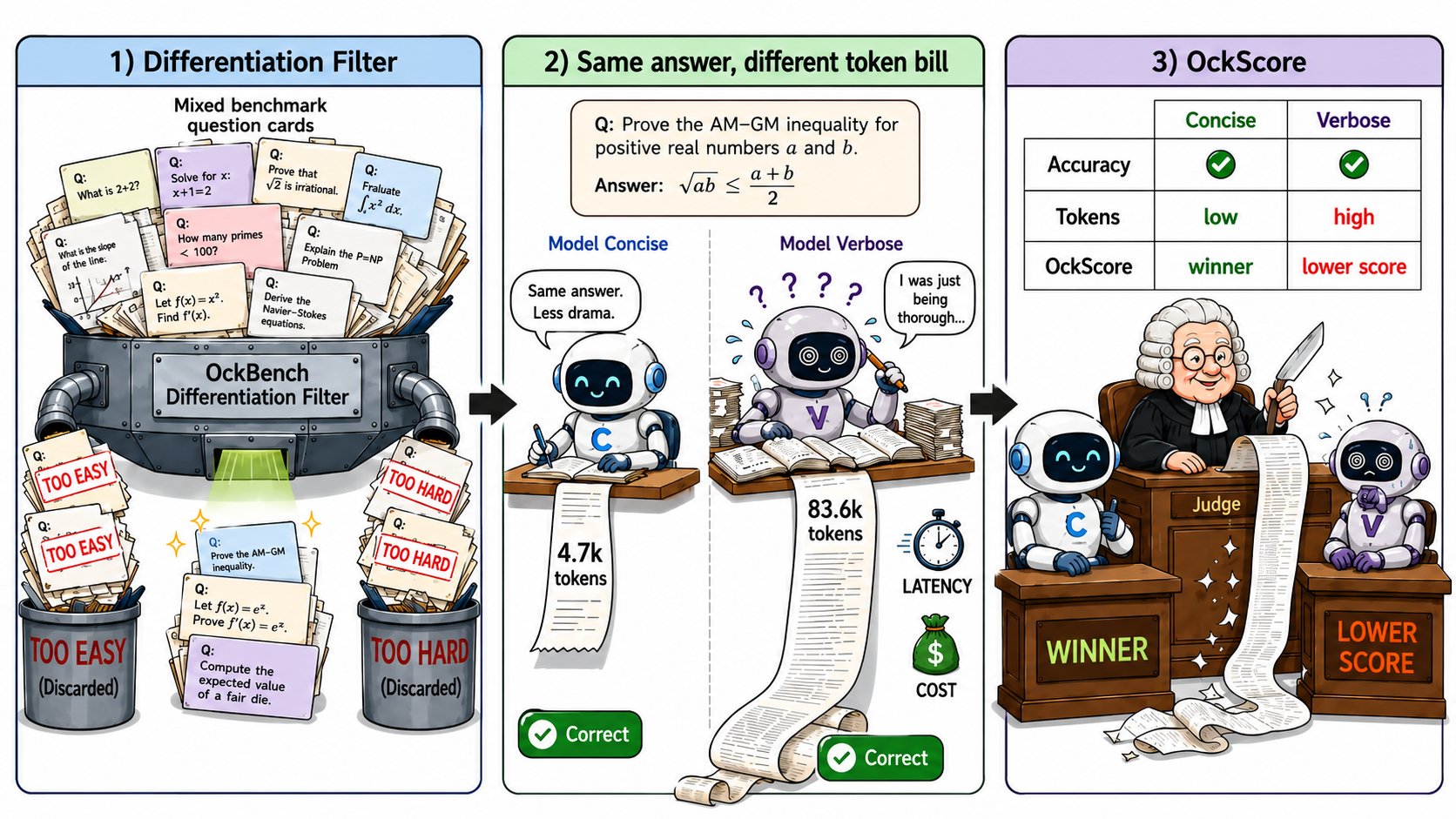}
    \caption{\textbf{OckBench overview.} OckBench filters for questions that expose token-efficiency gaps, compares models that reach the same answer with different token budgets, and uses OckScore to reward concise, correct reasoning.}
    \label{fig:teaser}
\end{figure}
}

\section{Introduction}

\begin{quote}
\textit{``Entities must not be multiplied beyond necessity.''}
\begin{flushright}
The Principle of Ockham's Razor
\end{flushright}
\end{quote}

State-of-the-art large language models (LLMs), such as GPT-5.5 \citep{openai2026gpt55} and Claude Opus 4.7 \citep{anthropic2026claudeopus}, increasingly use long reasoning traces to solve hard tasks. This improves accuracy, but also creates a token-efficiency bottleneck. Solving six International Olympiad in Informatics problems can now take over ten hours \citep{openai_learning_to_reason_2024}, turning inference into a long decoding marathon \citep{imojourney}. The same issue appears in agentic applications: systems such as Claude Code \citep{anthropic2026claudecode}, Codex \citep{openai2026codex}, and OpenClaw \citep{openclaw2026docs} are built for long-running tool-use workflows, where token budgets compound across steps. Every extra generated token, including reasoning before the final answer, adds latency, cost, and inference-time energy use. Recent work shows that LLM inference energy is strongly tied to output length, so long generations can accumulate large energy and carbon costs at deployment scale \citep{poddar2025sustainablenlpinsightsbenchmarking,fernandez2025energyconsiderationslargelanguage,jegham2025hungryaibenchmarkingenergy}.

Current benchmarks, such as HELM \citep{liang2023holistic} and Chatbot Arena \citep{chiang2024chatbotarena}, mainly measure \textit{output quality}. They say little about the token bill. In practice, many models spend far more tokens than needed. Even within one frontier family, similar accuracy can come with very different token budgets: Claude Opus-4.6 averages \textbf{28.6k} tokens, while Claude Opus-4.7 averages \textbf{7.5k} at nearly the same accuracy (Figure~\ref{fig:efficiency_gap_7b}). This makes token efficiency a real system bottleneck.


This variance is not an isolated anomaly. Our analysis shows a systematic token-efficiency gap between open-weight and proprietary models (Figure~\ref{fig:bubble_plot}). While the strongest open-weight models can approach proprietary accuracy, they often need many more tokens. For example, DeepSeek-V4-Flash (max) reaches \textbf{82.5\%} overall accuracy, comparable to GPT-5.4 (medium) at \textbf{82.0\%}, but consumes \textbf{26.3$\times$} more tokens (Figure~\ref{fig:efficiency_gap_open_closed}). We also find that token efficiency scales with model capacity. Across the Qwen3.5 family, scaling from 9B to 397B-A17B improves accuracy from \textbf{21.5\%} to \textbf{67.5\%} while reducing average output length by \textbf{4.4$\times$}. Smaller models often pay an ``Overthinking Tax'': they reason at length, spend more tokens, and can become more expensive in real deployment (Figure~\ref{fig:efficiency_gap_scaling}).

These compounding observations expose a blind spot in current evaluation. \textbf{Correctness alone is no longer enough}: two models can reach the same answer, while one takes the scenic route through a small novel. Token efficiency is therefore part of model intelligence and should be evaluated alongside accuracy.
Recent work studies reasoning efficiency from related angles~\citep{li2025thinkbenchevaluatingthinkingefficiency,pu2025thoughtterminator,weston2025optimalthinkingbenchevaluatingunderthinking,artificialanalysis2026}. However, these efforts do not center on a deployment-facing, hardware-agnostic benchmark that isolates cross-model token-efficiency gaps on reasoning tasks with targeted item selection and a unified accuracy--token score.

\afterpage{\begin{figure*}[!t]
    \centering
    \begin{subfigure}{0.32\textwidth}
        \centering
        \includegraphics[width=\linewidth]{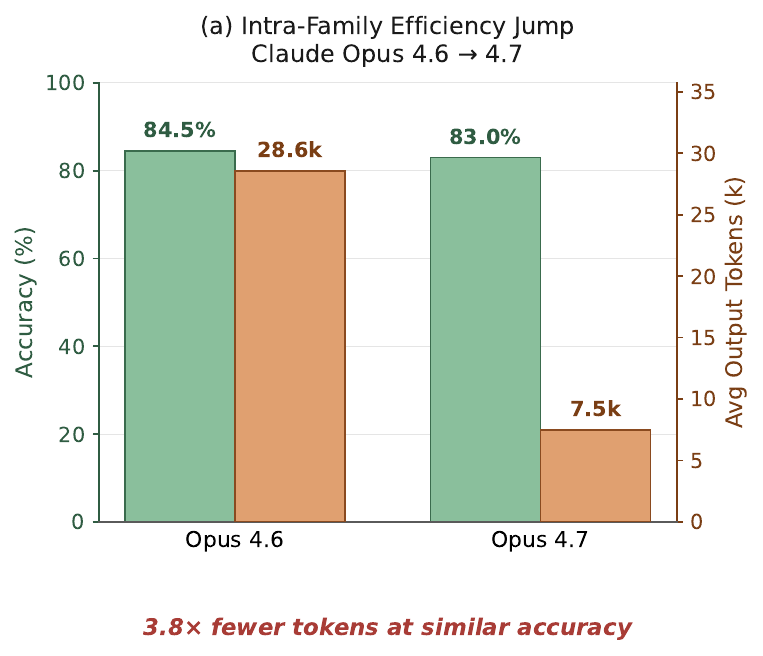}
        \caption{Intra-family efficiency jump}
        \label{fig:efficiency_gap_7b}
    \end{subfigure}
    \hfill
    \begin{subfigure}{0.32\textwidth}
        \centering
        \includegraphics[width=\linewidth]{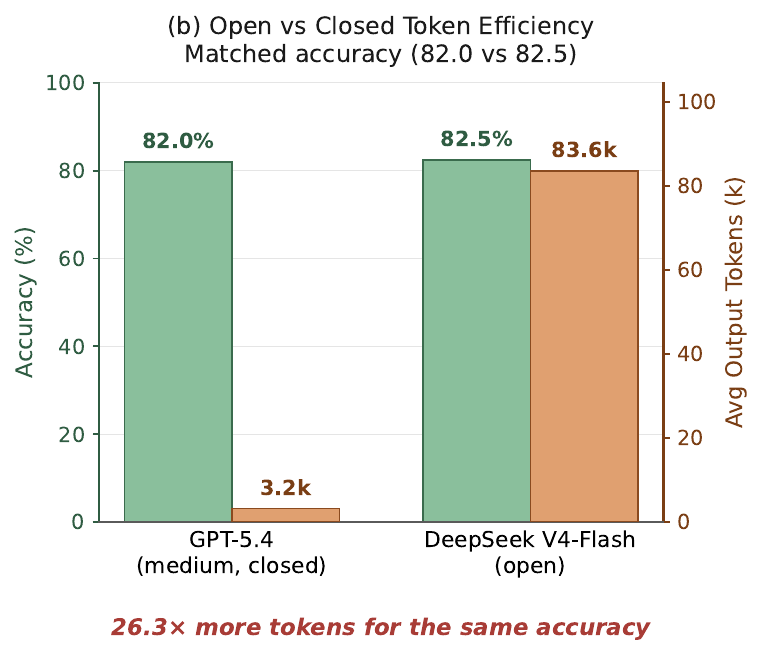}
        \caption{Open vs. closed efficiency}
        \label{fig:efficiency_gap_open_closed}
    \end{subfigure}
    \hfill
    \begin{subfigure}{0.32\textwidth}
        \centering
        \includegraphics[width=\linewidth]{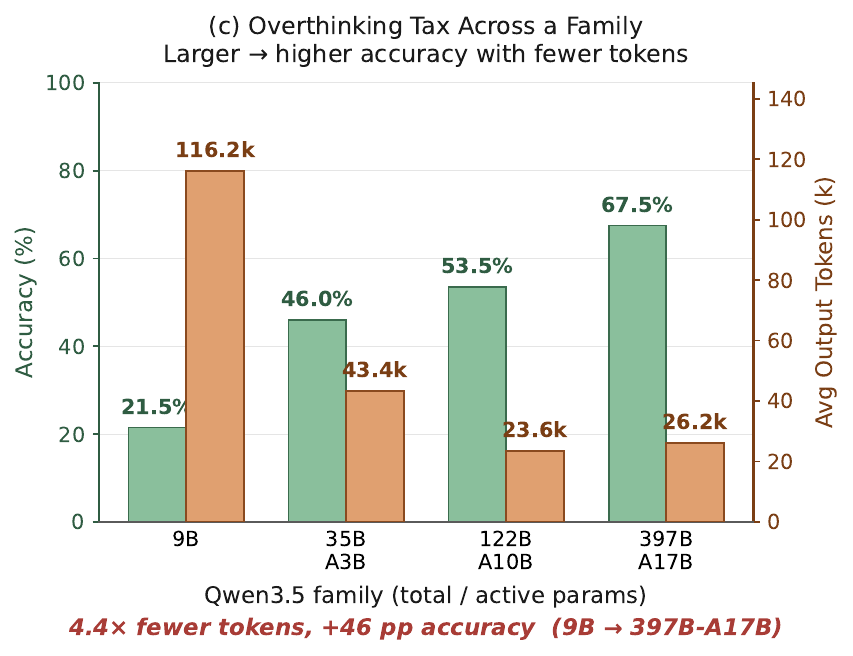}
        \caption{Overthinking tax across Qwen}
        \label{fig:efficiency_gap_scaling}
    \end{subfigure}
    \caption{\textbf{Efficiency Gaps in Reasoning Models.} (a) Within the Claude Opus family, Opus 4.7 preserves nearly the same accuracy as Opus 4.6 while reducing average output length from 28.6k to 7.5k tokens. (b) At matched accuracy, DeepSeek V4-Flash (max) reaches 82.5\% but uses 83.6k tokens on average, a 26.3$\times$ token gap compared with 3.2k for GPT-5.4 (medium). (c) Across the Qwen3.5 family, scaling from 9B to 397B-A17B improves accuracy from 21.5\% to 67.5\% while reducing output length by 4.4$\times$, illustrating that smaller models can pay a large overthinking tax.}
    \label{fig:efficiency_gap_main}
\end{figure*}
}


To address this gap, we introduce OckBench, a benchmark that jointly measures accuracy and token consumption to evaluate model reasoning intelligence. Through this framework, we formalize the concept of \textit{Per-Token Intelligence}, a critical dimension that defines a model's ability to solve complex problems correctly and concisely.

OckBench covers three core domains: math, coding, and science. It draws from reasoning-heavy datasets such as AIME, MBPP \citep{austin2021program}, and GPQA \citep{rein2023gpqagraduatelevelgoogleproofqa}. To reveal efficiency gaps, OckBench selects problems with diverse reasoning paths and naturally wide variation in token use. These problems separate models that solve tasks concisely from models that babble their way to the answer. We also design OckScore ($S_{Ock}$), a unified metric that prioritizes correctness while logarithmically penalizing unnecessary verbosity. Figure~\ref{fig:teaser} summarizes this pipeline: filter for efficiency-revealing questions, compare token budgets under matched correctness, and score concise reasoning higher.

Through evaluations on OckBench, we uncover several lessons for model development. The large token-efficiency gap between open-weight and proprietary models suggests that future training should optimize reasoning density, not only final accuracy. We also show that smaller models are not always cheaper: lower per-token cost can be offset by worse token efficiency. Our main contributions are:



\begin{itemize}
    \item \textbf{Per-Token Intelligence:} We introduce token efficiency as a core dimension of LLM evaluation. A stronger model should be accurate and avoid treating tokens as free.
    \item \textbf{OckBench and OckScore:} We present an open, hardware-agnostic benchmark that jointly measures accuracy and token efficiency. With a ``Differentiation Filter,'' OckBench selects tasks that expose efficiency gaps. OckScore ($S_{Ock}$) then penalizes unnecessary verbosity while preserving correctness as the first priority.
    \item \textbf{Empirical Insights and the ``Overthinking Tax'':} We evaluate a broad set of models and quantify a clear efficiency gap between open-weight and proprietary models. We also define the ``Overthinking Tax,'' where verbose smaller models can cost more in deployment.
    \item \textbf{Optimization Pathways:} We show that reasoning efficiency can be improved. With training-free model interpolation and difficulty-aware reinforcement learning, models can ``Ockhamize'' reasoning and improve OckScore.
\end{itemize}







\section{Background \& Related Work}

\subsection{The Blind Spot in Current Evaluations}
Current benchmarks typically focus on output quality or system performance, leaving intrinsic token efficiency undermeasured.

\textbf{Accuracy-Centric Benchmarks} suites such as HELM \citep{liang2023holistic}, LM-Eval \citep{eval-harness}, and Chatbot Arena \citep{chiang2024chatbotarena} measure correctness or preference, yet they do not penalize verbose reasoning and can even reward it through length bias. \textbf{System-Centric Benchmarks} such as MLPerf \citep{reddi2019mlperf} measure latency and throughput, but these reflect serving efficiency, not whether the model itself reasons concisely.

\textbf{Reasoning efficiency} has attracted more attention in recent work. Think-Bench~\citep{li2025thinkbenchevaluatingthinkingefficiency} studies process quality and thinking efficiency, ThoughtTerminator~\citep{pu2025thoughtterminator} analyzes overthinking with difficulty-aware token-spend analysis, and OptimalThinkingBench~\citep{weston2025optimalthinkingbenchevaluatingunderthinking} evaluates adaptive compute allocation across easy and hard tasks. Related optimization work, including DART~\citep{zhang2025dart} and ConciseR~\citep{song2025conciser}, further shows that reasoning length can be adapted or trained down while preserving accuracy. These works mainly ask how much compute a model should spend as task hardness changes, or how to make one model more concise. OckBench targets a different question: when models have similar accuracy, how much do they diverge in output-token efficiency? Artificial Analysis~\citep{artificialanalysis2026} also reports token usage and cost at the platform level. OckBench adds efficiency-revealing questions, a unified accuracy--token metric, and a local implementation.

\begin{figure*}[t]
    \centering
    \includegraphics[width=\linewidth]{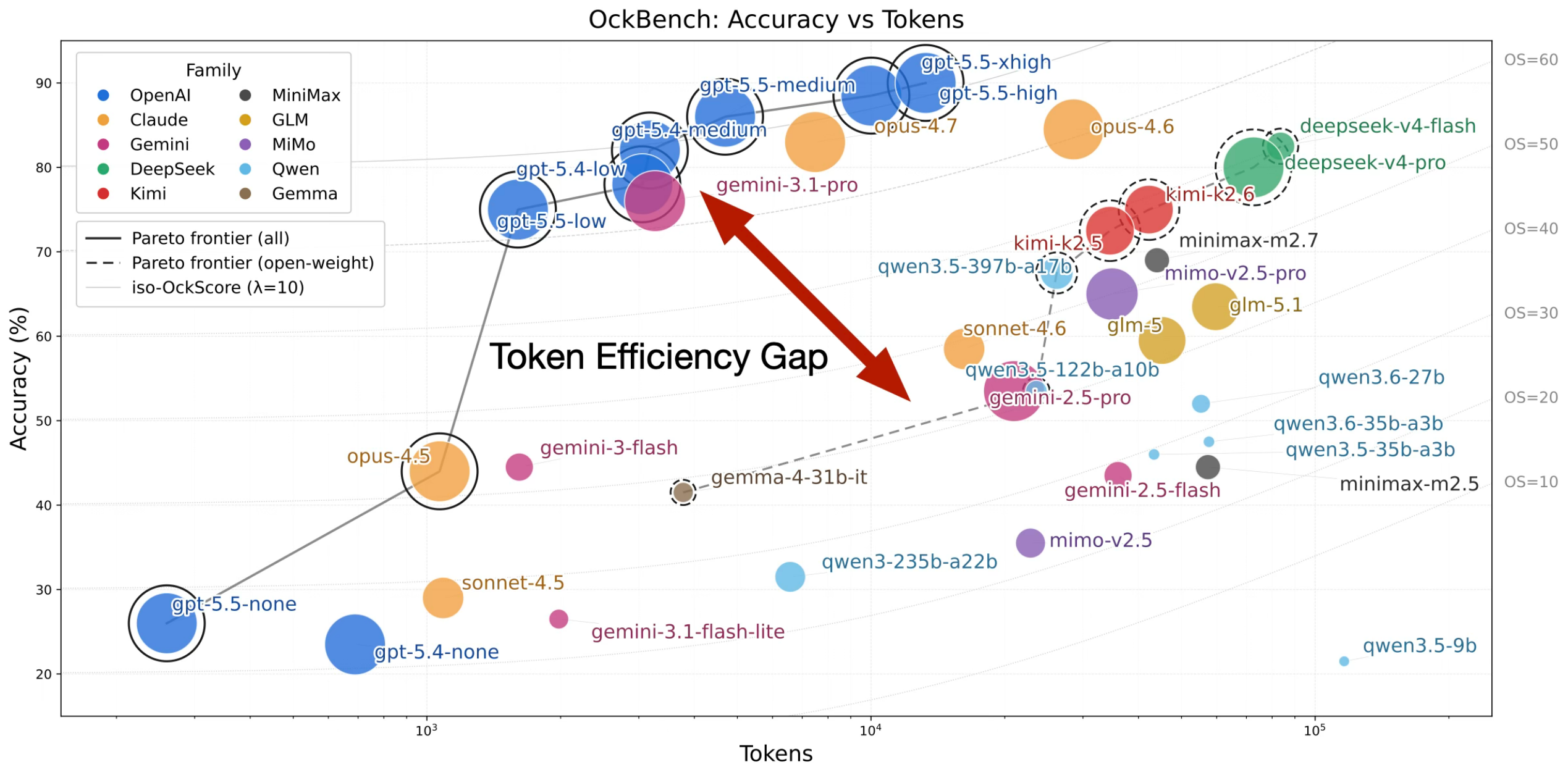}
    \caption{\textbf{OckBench Accuracy--Token Trade-off.} Each point is a model or effort setting in the 37-setting evaluation, plotted by overall accuracy and average full-output tokens (reasoning plus answer) on a log scale; bubble area denotes the model-size proxy. The dashed frontier marks non-dominated accuracy--token trade-offs, and the gray iso-OckScore contours use the default $\lambda=10$ calibration.}
    \label{fig:bubble_plot}
    \vspace{0.2em}
    \begin{minipage}{0.97\linewidth}
        \footnotesize \emph{Size proxy.} For closed-source models without public parameter counts, models in the same tier are assigned the same size proxy (e.g., Claude Opus and Gemini Pro; Claude Sonnet and Gemini Flash). For open-weight dense models, we use the released parameter count. For MoE models, we use $\sqrt{N_{\mathrm{total}}N_{\mathrm{active}}}$.
    \end{minipage}
\end{figure*}

\subsection{The Rising Cost of Reasoning and Token Efficiency Optimization}
Inference-time scaling has made long reasoning traces common in both proprietary and open-weight models. Methods such as Chain-of-Thought \citep{wei2023chainofthoughtpromptingelicitsreasoning}, Best-of-$N$ \citep{stiennon2020learning}, and search- or RL-based decoding \citep{yao2023tree, zuo2025ttrl} often improve difficult-task accuracy, but they also increase output length substantially \citep{jaech2024openai, guo2025deepseek}. This growth is expensive: Epoch AI reports that reasoning-model output lengths are increasing by roughly $5\times$ per year \citep{epoch2025outputlength}, while latency, compute, and API pricing all scale with token count \citep{saleem_wccftech_top30_2025}.

Existing efficiency work has mostly targeted systems and compression, such as FlashAttention \citep{dao2022flashattentionfastmemoryefficientexact} and KV-cache quantization \citep{lin2025qservew4a8kv4quantizationcodesign, kang2024gearefficientkvcache}, which improve serving but do not reduce the number of reasoning tokens a model chooses to generate. Newer methods, including DIET \citep{chen2025overthinker}, model interpolation \citep{wu2025revisiting}, DART~\citep{zhang2025dart}, ConciseR~\citep{song2025conciser}, and ThoughtTerminator \citep{pu2025thoughtterminator}, directly reduce redundant reasoning while preserving accuracy. Unit-test-based evaluation such as UTMath~\citep{yang2024utmath} is complementary to this direction, emphasizing executable verification for mathematical reasoning. OckBench makes output-token cost visible through a benchmark and per-token score.

\section{Benchmark Design: OckBench}
\label{sec:benchmark_design}

OckBench extends reasoning evaluation from accuracy alone to accuracy together with \textit{token efficiency}. Instead of ranking models only by correctness, it selects tasks that expose clear differences in token use. This section describes the domains, item-selection pipeline, and evaluation protocol.

\subsection{Benchmark Composition}
To ensure broad coverage of reasoning modalities, OckBench spans three complementary domains: mathematics, software engineering, and scientific reasoning.

\paragraph{Mathematics and Reasoning.}
Mathematical problem-solving serves as the core testbed for logical rigor. We construct our pool from {GSM8K}~\citep{cobbe2021trainingverifierssolvemath}, {AIME 2024/2025}~\citep{maa_aime_page}, {OlympiadBench}~\citep{he2024olympiadbenchchallengingbenchmarkpromoting}, and {MATH500}~\citep{lightman2023letsverifystepstep}. To probe frontier reasoning beyond current saturation points, we also include the math subset of {Humanity's Last Exam}~\citep{phan2025humanitysexam} and {AMO-Bench}~\citep{an2025amobenchlargelanguagemodels}, spanning grade-school arithmetic to competition-level number theory.

\paragraph{Software Engineering.}
Code generation proxies real-world logical synthesis and planning. We use lightweight {MBPP}~\citep{austin2021program} and {LiveCodeBench}~\citep{jain2024livecodebench} variants. 

\paragraph{Scientific Reasoning.}
We include {ScienceQA}~\citep{lu2022learn}, STEM {MMLU}~\citep{hendrycks2021measuringmassivemultitasklanguage}, and {GPQA-Diamond}~\citep{rein2023gpqagraduatelevelgoogleproofqa} to test knowledge-constrained reasoning under technical load.

\subsection{Question Selection Strategy: The Differentiation Filter}
\label{subsec:selection_strategy}

Existing datasets were not designed to measure token efficiency, and random sampling is therefore often uninformative. OckBench instead applies a \textit{Differentiation Filter} that selects problems on which models exhibit substantial differences in token usage.

Formally, let $\mathcal{M}$ denote a diverse set of representative models and $\mathcal{D}_{pool}$ the initial candidate pool. For each problem $x \in \mathcal{D}_{pool}$, we collect the output-token lengths $\mathcal{L}_x = \{ \text{len}(m(x)) \mid m \in \mathcal{M} \}$. We then construct the final benchmark set $\mathcal{D}_{ock}$ using two criteria:

\begin{enumerate}
    \item \textbf{Difficulty Banding:} We filter for problems where the average accuracy across $\mathcal{M}$ falls within a target band of $0.1 \leq \text{acc}(x) \leq 0.9$. This removes trivial instances and intractable queries, focusing on the ``reasoning frontier'' where efficiency matters most.
    \item \textbf{Maximizing \#Tokens Variance:} From the remaining pool, we select the top-$k$ instances that maximize the variance of token consumption, $\text{Var}(\mathcal{L}_x)$. Specifically, we determine $k$ proportionally from each domain's candidate pool, yielding 200 problems: 100 for mathematics, 60 for software engineering, and 40 for scientific reasoning.
\end{enumerate}

\textbf{Rationale.} High variance in token consumption implies that a problem admits multiple valid reasoning paths, with some efficient and others convoluted. By selecting these instances, OckBench clearly exposes the \textit{Efficiency Gap} between a model that ``Ockhamizes'' its reasoning and one that ``babbles.'' This ensures that the benchmark penalizes unnecessary verbosity without penalizing the necessary complexity of hard problems. The 200-item set used in this paper is treated as the fixed OckBench v1.0 split; future refreshes should be released as new versions.

\subsection{Evaluation Protocol}
We use greedy decoding (temperature $= 0$) to avoid prompt variation and sampling noise. For mathematics and science, we apply rule-based answer extraction and exact matching; for coding, we use functional execution against unit tests. Our efficiency metric is \textit{Output Tokens}, the full generated response before EOS, including both intermediate reasoning and the final answer, not just answer tokens. This is the token stream exposed by the evaluated endpoint or local tokenizer, so OckBench measures the provider-observable generated-token budget under documented serving settings. Because latency and billing scale with total generated tokens, this metric reflects deployment cost. A tokenizer sanity check applying mainstream tokenizers to a reference model's complete OckBench output finds at most 10\% divergence (Appendix~\ref{sec:tokenizer_sanity_check}); the large efficiency gaps below are therefore not dominated by tokenizer artifacts.





\subsection{Unified OckScore}
To evaluate accuracy and token efficiency jointly, we introduce \textit{OckScore} ($S_{\text{ock}}$). In deployment, correctness and token cost are coupled: a concise correct answer is preferable to a substantially longer correct trace. Rather than splitting evaluation by difficulty level~\citep{weston2025optimalthinkingbenchevaluatingunderthinking}, OckBench applies a single token-penalized accuracy score across differentiation-filtered items, asking how concise a model is when solving reasoning-level problems.


Our preference ordering is:
\begin{equation*}
    \langle \textsc{Correct}, \textsc{Short}\rangle \succ \langle \textsc{Correct}, \textsc{Long}\rangle 
    \succ \langle \textsc{Wrong}, \textsc{Short}\rangle \succ \langle \textsc{Wrong}, \textsc{Long}\rangle
\end{equation*}
We quantify it with a penalized accuracy metric:
\begin{equation*}
    S_{\text{ock}} = \text{Accuracy} - \lambda \cdot \log\left(\frac{T}{C} + 1\right)
\end{equation*}
where $\text{Accuracy}$ is the percentage pass rate (0-100), $T$ is the average number of output tokens, $\lambda$ is a penalty coefficient, and $C$ is a normalization constant. In our experiments, we set $\lambda=10$ and $C=10,000$. This formulation is motivated by three considerations:

\textbf{Accuracy Prioritization.}
Correctness is the primary utility of a reasoning model. Using accuracy as the base term and subtracting a logarithmic penalty treats efficiency as a cost, not a multiplier, while keeping correct solutions above concise failures.

\textbf{Logarithmic Scaling of Efficiency.}
We use a logarithmic penalty on normalized token length ($\log(1+T/C)$) because reasoning lengths vary by orders of magnitude. A linear penalty would underweight short traces or over-penalize long but necessary traces; the logarithmic form compresses this range while still penalizing unnecessary verbosity.

\textbf{Calibration with Model Priors.}
The parameters $\lambda=10$ and $C=10{,}000$ avoid rewarding concise but inaccurate models. Empirically, this calibration ranks stronger frontier models and larger same-family models above weaker alternatives while reflecting accuracy--brevity trade-offs: GPT-5.5 (medium) outranks GPT-5.4 (medium), Gemini-3.1-Pro outranks Gemini-2.5-Pro, and Qwen3.5-397B-A17B outranks the much more verbose Qwen3.5-9B. It also penalizes the ``overthinking'' tax described in Section \ref{sec:analysis}; Appendix~\ref{subsec:interpreting_ockscore} gives the full rationale and nearby-$\lambda$ stability analysis.

\section{Experiments}
\label{sec:experiments}

\subsection{Experimental Setup and Protocol}
\label{sec:setup}

We follow the benchmark composition, item selection, and evaluation protocol in Section~\ref{sec:benchmark_design}. Specifically, we evaluate all models on the fixed 200-item OckBench v1.0 split selected by the Differentiation Filter, and use greedy decoding (temperature $= 0$) to ensure statistical stability. The supplementary material releases the anonymized v1.0 selected subset, prompt/config templates, generation caps, reasoning-effort labels, evaluation scripts, and per-setting result artifacts used to audit the leaderboard.

\subsection{Models}
\label{sec:models}
We evaluate state-of-the-art proprietary and open-weight systems across provider-side reasoning-effort settings. The evaluation contains 37 model/setting entries; labels such as \textit{low}, \textit{medium}, \textit{high}, and \textit{xhigh} denote provider-controlled reasoning effort, while \textit{none} denotes the default no-effort setting.

\begin{itemize}
    \item \textbf{Commercial Models:} We evaluate OpenAI's GPT-5.4 and GPT-5.5 families across no-effort and low/medium/high/xhigh effort settings \citep{openai_models_2026}; Anthropic's Claude Opus 4.5/4.6/4.7 and Claude Sonnet 4.5/4.6 variants \citep{anthropic_model_system_cards_2026}; and Google's Gemini 2.5, 3, and 3.1 Pro/Flash/Flash-Lite settings \citep{gemini25family2025,gemini3series2026,gemini31pro2026}.
    \item \textbf{Open-Weight Models:} We evaluate DeepSeek-V4-Flash and DeepSeek-V4-Pro \citep{deepseek_v4_hf_2026}; Kimi-K2.5 and Kimi-K2.6 \citep{kimiteam2026kimik25visualagentic,kimi_k26_docs_2026}; MiniMax-M2.5 and MiniMax-M2.7 \citep{minimax_m25_2026,minimax_m27_2026}; GLM-5 \citep{glm5_arxiv_2026} and GLM-5.1 \citep{glm51_docs_2026}; MiMo-v2.5 and MiMo-v2.5-Pro \citep{xiaomi_mimo_2026}; Gemma-4-31B-IT \citep{google_gemma4_2026}; and Qwen3.5/3.6 plus Qwen3-235B-A23B variants \citep{qwen35_2026,qwen36_2026,qwen3technicalreport}.
\end{itemize}


\subsection{Main Results}
\label{sec:main_result}

\begin{table*}[t]
    \centering
    \setlength{\tabcolsep}{5pt}
    \renewcommand{\arraystretch}{1.2}
    \footnotesize
    \caption{\textbf{OckBench Leaderboard.} Selected variants from each model family, ranked by OckScore ($S_{Ock}$). The table reports accuracy, average full-output tokens, and total output cost for the 200-item OckBench run (USD) from OpenRouter completion prices~\citep{openrouter_pricing_2026}. The full leaderboard appears in Appendix~\ref{app:full_leaderboard}. Shaded rows denote open-weight models.}
    \label{tab:leaderboard}
    \resizebox{\textwidth}{!}{
    \begin{tabular}{llrrrrrrrrrr}
    \toprule
    \multirow{2}{*}{\textbf{Model}} & \multirow{2}{*}{\textbf{Org}}
    & \multicolumn{2}{c}{\textbf{Coding}}
    & \multicolumn{2}{c}{\textbf{Science}}
    & \multicolumn{2}{c}{\textbf{Math}}
    & \multicolumn{2}{c}{\textbf{Overall}}
    & \multirow{2}{*}{\textbf{Cost}}
    & \multirow{2}{*}{\textbf{OckScore}} \\
    \cmidrule(lr){3-4} \cmidrule(lr){5-6} \cmidrule(lr){7-8} \cmidrule(lr){9-10}
    & & \textbf{Acc.} & \textbf{Tok.} & \textbf{Acc.} & \textbf{Tok.} & \textbf{Acc.} & \textbf{Tok.} & \textbf{Acc.} & \textbf{Tok.} & & \\
    \midrule
    GPT-5.5 (medium) & OpenAI & 96.7 & 1,739 & 85.0 & 4,043 & 80.0 & 6,723 & 86.0 & 4,692 & 28.2 & 82.2 \\
    GPT-5.5 (xhigh) & OpenAI & 100.0 & 4,501 & 87.5 & 11,962 & 85.0 & 19,115 & 90.0 & 13,271 & 79.6 & 81.6 \\
    GPT-5.4 (medium) & OpenAI & 93.3 & 2,338 & 75.0 & 4,308 & 78.0 & 3,235 & 82.0 & 3,177 & 9.5 & 79.2 \\
    Claude-Opus-4.7 & Anthropic & 85.0 & 1,878 & 85.0 & 6,076 & 81.0 & 11,406 & 83.0 & 7,481 & 37.4 & 77.4 \\
    Gemini-3.1-Pro & Gemini & 86.7 & 2,723 & 70.0 & 1,934 & 72.0 & 4,130 & 76.0 & 3,263 & 7.8 & 73.2 \\
    Claude-Opus-4.6 & Anthropic & 88.3 & 7,187 & 70.0 & 27,217 & 88.0 & 41,964 & 84.5 & 28,582 & 142.9 & 71.0 \\
    \rowcolor{black!10} DeepSeek-v4-flash (max) & DeepSeek & 96.7 & 32,709 & 75.0 & 90,834 & 77.0 & 111,211 & 82.5 & 83,585 & 4.7 & 60.1 \\
    \rowcolor{black!10} Kimi-K2.6 & Kimi & 68.3 & 24,408 & 75.0 & 41,898 & 79.0 & 53,082 & 75.0 & 42,243 & 29.6 & 58.5 \\
    \rowcolor{black!10} Qwen3.5-397B-A17B & Qwen & 85.0 & 19,404 & 67.5 & 12,355 & 57.0 & 35,771 & 67.5 & 26,178 & 12.3 & 54.6 \\
    \rowcolor{black!10} MiniMax-M2.7 & MiniMax & 85.0 & 26,207 & 70.0 & 28,831 & 59.0 & 60,806 & 69.0 & 44,031 & 10.6 & 52.1 \\
    \rowcolor{black!10} MiMo-v2.5-Pro & MiMo & 75.0 & 14,569 & 60.0 & 30,384 & 61.0 & 48,868 & 65.0 & 34,882 & 20.9 & 50.0 \\
    Claude-Sonnet-4.6 & Anthropic & 88.3 & 12,471 & 45.0 & 27,366 & 46.0 & 13,973 & 58.5 & 16,201 & 48.6 & 48.9 \\
    \rowcolor{black!10} GLM-5.1 & GLM & 70.0 & 58,157 & 67.5 & 40,290 & 58.0 & 68,272 & 63.5 & 59,641 & 41.7 & 44.1 \\
    Gemini-3-Flash & Gemini & 55.0 & 2,088 & 57.5 & 2,343 & 33.0 & 1,076 & 44.5 & 1,615 & 1.0 & 43.0 \\
    \rowcolor{black!10} Gemma-4-31B-IT & Gemma & 51.7 & 4,374 & 45.0 & 3,118 & 34.0 & 3,681 & 41.5 & 3,776 & 0.3 & 38.3 \\
    \bottomrule
    \end{tabular}
    }
\end{table*}

Table~\ref{tab:leaderboard} reports the selected main-text leaderboard ranked by OckScore ($S_{\text{ock}}$), with an output-only total-cost estimate for the 200-item run from current OpenRouter completion prices. The full leaderboard is in Appendix~\ref{app:full_leaderboard}. Figure~\ref{fig:bubble_plot} shows the accuracy--token trade-off and encodes the model-size proxy by bubble area.

\paragraph{Accuracy and efficiency do not align.}
\textbf{GPT-5.5 (medium)} ranks first by OckScore, with 86.0\% accuracy and 4,692 average tokens ($S_{\text{ock}}=82.2$). Increasing test-time effort improves raw accuracy but not necessarily efficiency: GPT-5.5 (xhigh) reaches 90.0\% accuracy, yet uses 13,271 tokens, about \textbf{2.8$\times$} more, costs \$79.6 instead of \$28.2 for the full run, and scores slightly lower ($81.6$). This gap shows why accuracy alone is incomplete.

\paragraph{Frontier models differ sharply in reasoning density.}
\textbf{GPT-5.4 (medium)} remains highly efficient, achieving 82.0\% accuracy with 3,177 average tokens ($S_{\text{ock}}=79.2$). Claude Opus-4.7 reaches similar accuracy (83.0\%) but uses 7,481 tokens, while Claude Opus-4.6 reaches 84.5\% accuracy at \textbf{28,582} tokens and drops to $S_{\text{ock}}=71.0$. Gemini-3.1-Pro offers another concise frontier point, with 76.0\% accuracy and 3,263 tokens. Figure~\ref{fig:bubble_plot} makes this spread visible: the best trade-offs lie on a narrow frontier, while many high-accuracy models are displaced rightward by long generations.

\paragraph{Open-weight models remain much more verbose.}
Among open-weight models, \textbf{DeepSeek-v4-flash (max)} is the strongest by OckScore ($60.1$), reaching 82.5\% accuracy. However, it uses 83,585 tokens, about \textbf{26.3$\times$} more than GPT-5.4 (medium) at nearly the same accuracy. The cost column shows why token efficiency still matters after pricing: despite a much lower per-token price, DeepSeek-v4-pro (max) is slightly less accurate than GPT-5.4 (medium) yet costs more for the full run (\$12.6 vs. \$9.5) because its outputs are much longer. Kimi-K2.6 also reaches competitive accuracy (75.0\%) but requires 42,243 tokens; compared with Gemini-3.1-Pro, it is slightly less accurate and about \textbf{12.9$\times$} more verbose. These results show that open-weight models are closing the accuracy gap, but still lag substantially in per-token reasoning density.

\paragraph{Summary.}
The leaderboard confirms the main goal of OckBench: models should be judged not only by correctness, but also by how efficiently they reach it. OckScore makes this trade-off visible, and these results show that reasoning efficiency remains a major separator across model classes.

\section{Analysis}
\label{sec:analysis}

\subsection{Token Distribution Analysis}
\label{sec:token_dist}

\begin{figure*}[t]
    \centering
    \includegraphics[width=\linewidth]{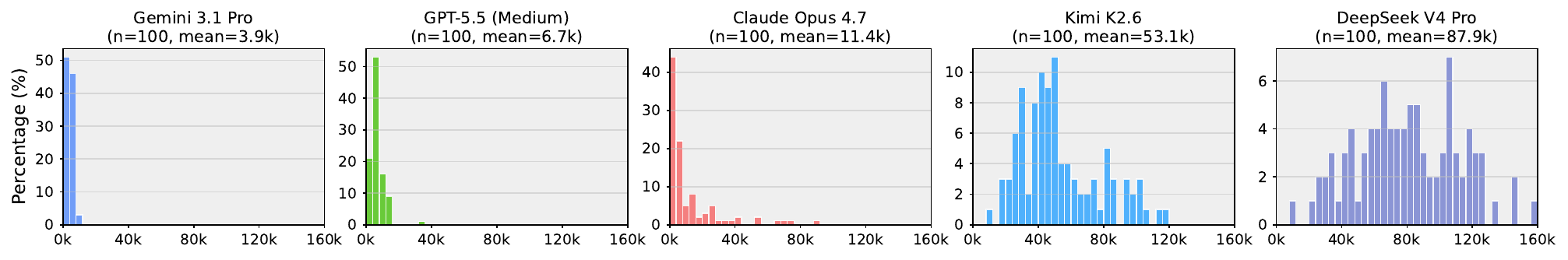}
    \caption{\textbf{Token Distributions on OckBench-Math.} Per-problem output-token histograms for five high-performing models. Frontier models concentrate in the low-thousand-token range, while Kimi K2.6 and DeepSeek V4 Pro (max) show heavier high-token tails.}
    \label{fig:token_histogram}
\end{figure*}

We first look beyond averages and examine how token use varies across individual math problems. Figure~\ref{fig:token_histogram} shows that frontier proprietary models such as Gemini-3.1-Pro, GPT-5.5, and Claude Opus-4.7 place much more mass in the \textbf{low-token range}, whereas Kimi-K2.6 and DeepSeek-V4-Pro (max) exhibit \textbf{heavy high-token tails}. This distributional view explains why average scores diverge even when accuracy is competitive: on math, Kimi-K2.6 averages 53.1k tokens and DeepSeek-V4-Pro (max) averages 87.9k, compared with 6.7k for GPT-5.5 (medium) and 4.1k for Gemini-3.1-Pro.

\begin{figure}[htbp]
    \centering
    \begin{subfigure}[b]{0.48\textwidth}
        \centering
        \includegraphics[width=\linewidth]{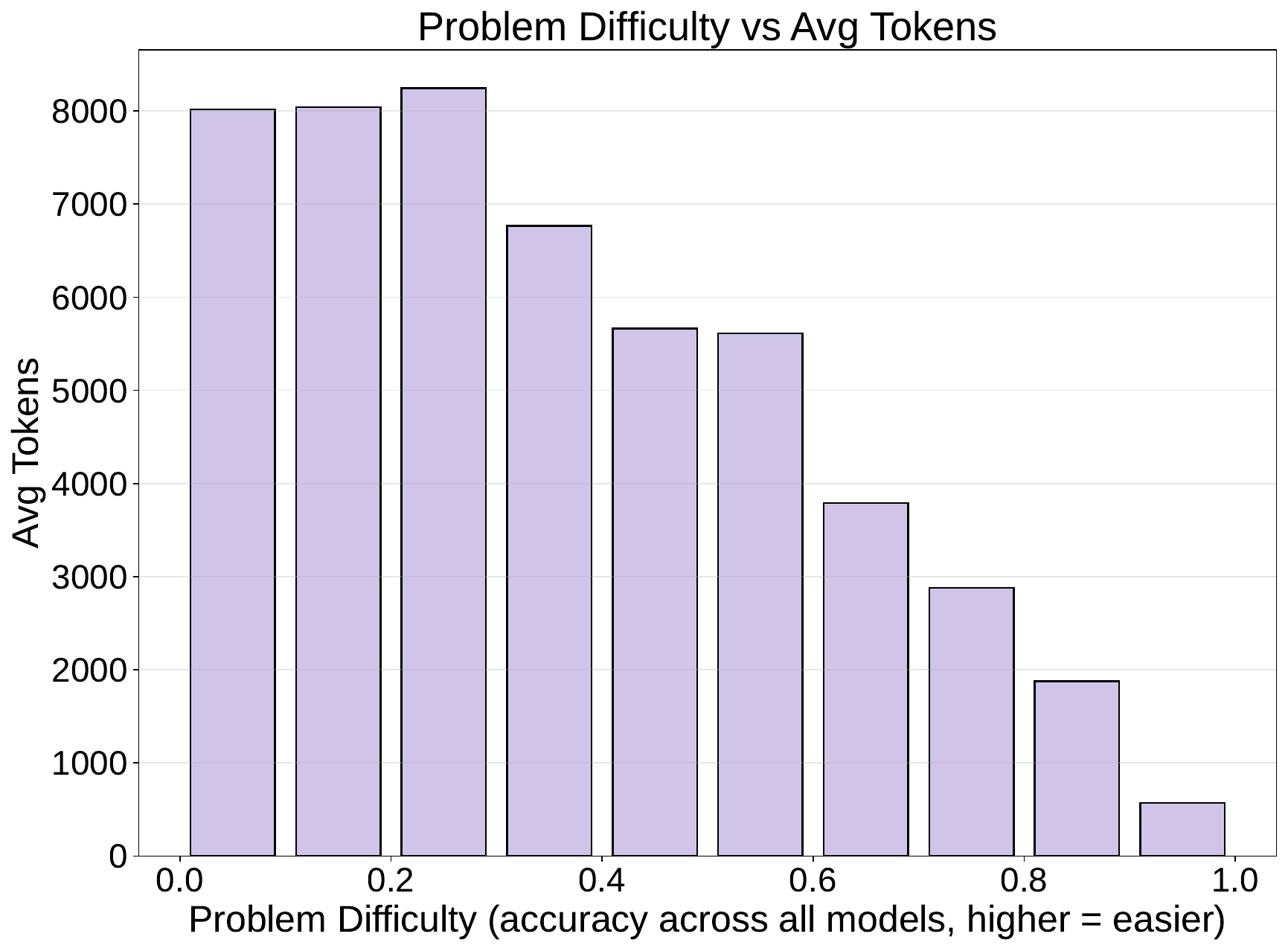}
        \caption{\textbf{Reasoning Cost vs. Difficulty.} Harder problems generally use more tokens; near-zero pass rates show overthinking tails.}
        \label{fig:difficulty_vs_tokens}
    \end{subfigure}
    \hfill 
    \begin{subfigure}[b]{0.48\textwidth}
        \centering
        \includegraphics[width=\linewidth]{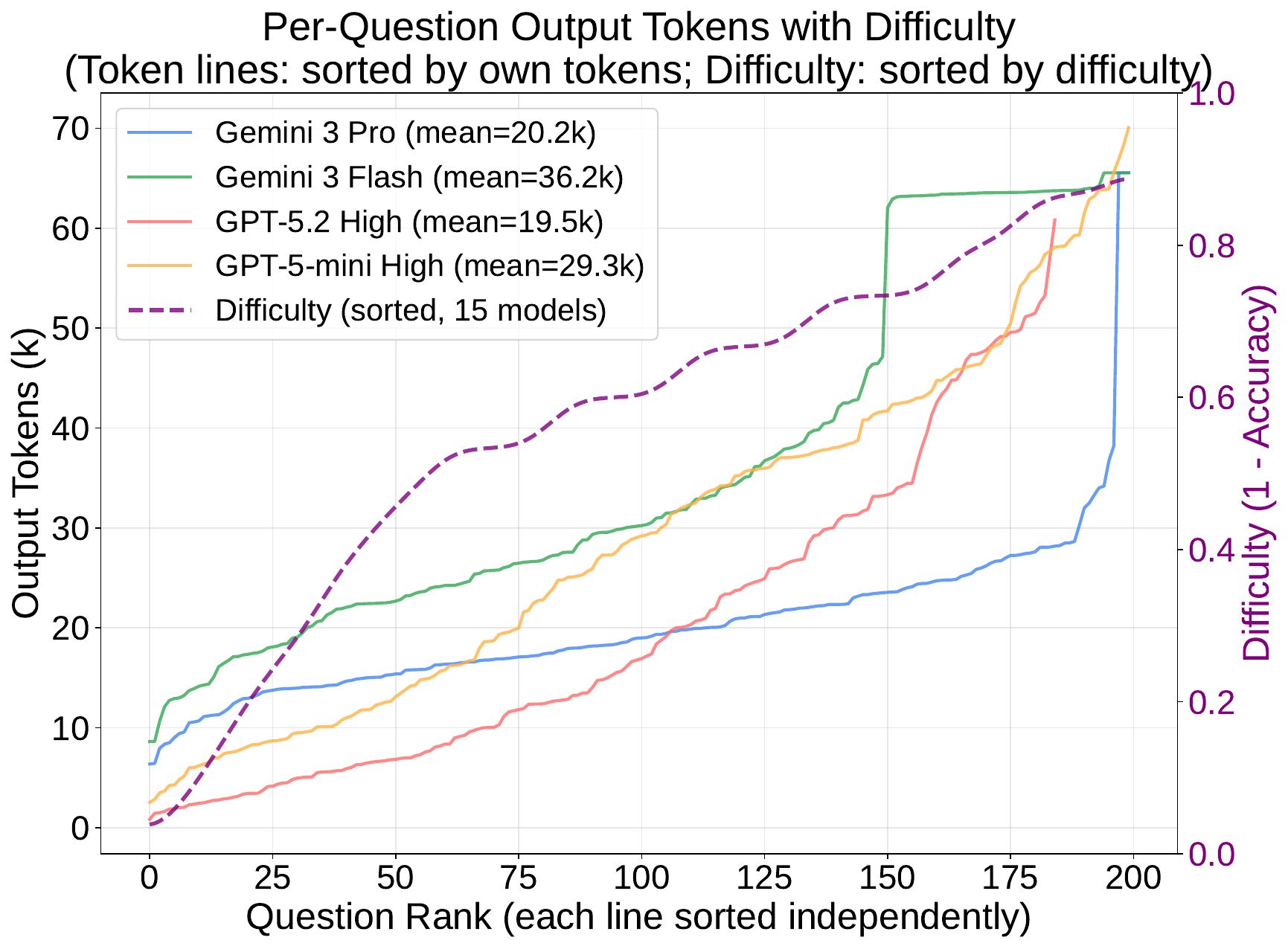}
        \caption{\textbf{Token Consumption Profiles.} Sorted traces show distinct GPT/Gemini scaling; the dashed line is smoothed difficulty.}
        \label{fig:token_profiles}
    \end{subfigure}
    
    \caption{Reasoning cost and token scaling on OckBench.}
\end{figure}

Figure~\ref{fig:token_profiles} compares GPT/Gemini token allocation across difficulty. Two strategies emerge:
\textit{Adaptive vs. Uniform Scaling.} The {GPT} series ({GPT-5.2} and {GPT-5-mini}) exhibits a \textit{difficulty-adaptive} trend. As difficulty increases (dashed purple line), token consumption scales linearly or super-linearly, allocating compute proportional to perceived complexity. In contrast, the {Gemini} family (e.g., {Gemini 3 Pro}) maintains a \textit{flat} profile across most of the benchmark (ranks 0--190), with stable \mbox{\textasciitilde}20k-token outputs across difficulty.

\paragraph{The ``Reasoning Cliff'' and Per-Token Intelligence.} 
This family-level contrast exposes a ``saturation cliff'' in the Gemini models. Unlike the GPT series' gradual ramp, Gemini token usage spikes abruptly to its generation limit, likely reflecting a context-window cap. In Figure~\ref{fig:token_profiles}, this appears as a distinct spike at the maximum token range ($\sim$64k) for Gemini 3 Flash.

This cliff serves as a proxy for model capacity. Gemini 3 Flash hits this saturation point much earlier (around rank 150 in Figure~\ref{fig:token_profiles}) than Pro (rank 195). We argue that delaying this cliff directly indicates \textit{per-token intelligence}: Pro resolves complex reasoning paths more succinctly than Flash, deferring the onset of ``reasoning collapse.'' This validates our hypothesis that efficiency is not merely about speed, but the information density of the reasoning chain.

\subsection{Tokens vs Difficulty}
Figure~\ref{fig:difficulty_vs_tokens} asks whether models spend more tokens on harder OckBench problems, using empirical pass rate as the difficulty proxy.
Across the 9 OpenAI reasoning settings, token use generally rises as pass rate falls, suggesting that models allocate more computation to harder items. The more interesting pattern appears near the low-accuracy end: for rarely solved problems, token consumption stays highly variable and often large instead of settling into a small failed-answer budget. This is a direct form of overthinking, paying latency for reasoning trajectories that still fail to converge.

\subsection{The ``Overthinking Tax'': When Efficient Models Pay More}
\label{sec:overthinking_tax}

This behavior becomes more costly in models assumed to be cheap. We call this the \textit{Overthinking Tax}: a model can look efficient by parameter count or cost-per-token, yet become expensive by generating long, low-yield reasoning traces. The Qwen3.5 family in Figure~\ref{fig:efficiency_gap_scaling} makes the effect concrete. \textbf{Qwen3.5-9B} reaches only 21.5\% overall accuracy while averaging 116.2k tokens, whereas Qwen3.5-397B-A17B reaches 67.5\% accuracy with 26.2k tokens. Scaling within the family therefore improves accuracy and reduces output length by \textbf{4.4$\times$}. Efficiency cannot be inferred from model size alone; without token-aware evaluation, a nominally smaller model can become the true deployment bottleneck.

\subsection{Efficiency Gap and Trends}
\label{sec:efficiency_trends}

Our analysis reveals two critical trends in recent reasoning models. Figure~\ref{fig:bubble_plot} gives the aggregate accuracy--token view, while Figure~\ref{fig:efficiency_gap_main} breaks the pattern down by model family and ownership type.

\paragraph{Open-closed efficiency gap.}
A distinct dichotomy remains between proprietary and open-weight models. Open-weight systems are closing the accuracy gap, but often remain far to the right of the proprietary frontier: at matched accuracy, DeepSeek-V4-Flash (max) reaches 82.5\% overall accuracy but uses \textbf{83.6k tokens}, compared with \textbf{3.2k tokens} for GPT-5.4 (medium) at 82.0\% accuracy. Kimi-K2.6 shows the same pattern at lower accuracy, requiring 42.2k tokens for 75.0\% accuracy. The gap is therefore about \textit{efficiency} as well as capability, requiring denser reasoning, not only scaling run-time.

\paragraph{Rapid convergence of frontier models.}
The frontier trajectory validates \textit{Per-Token Intelligence} as an optimization objective. \textbf{GPT-5.5 (medium)} achieves the best OckScore by balancing high accuracy with moderate token use, while GPT-5.5 (xhigh) attains higher raw accuracy but loses efficiency because it spends nearly three times as many tokens. The Claude Opus family shows a similar pattern: Opus-4.7 preserves nearly the same accuracy as Opus-4.6 while reducing average output length by \textbf{$4\times$}. Frontier progress is increasingly about moving upward and left in the accuracy--token plane, not merely increasing test-time compute.

\section{Methods to Improve Token Efficiency}
\label{sec:improvements}

We demonstrate how OckBench verifies efficiency-oriented optimization:
For the \textit{training-free} setting, we interpolate between \texttt{Qwen3-Thinking} and \texttt{Qwen3-Instruct} following prior model-interpolation work~\citep{wu2025revisiting}, using $\theta_{merged} = 0.6 \cdot \theta_{thinking} + 0.4 \cdot \theta_{instruct}$. The resulting \texttt{Qwen3-Mix-6:4} cuts average output length to \textbf{12,092} tokens, down from roughly 27k for the Thinking model, while retaining \textbf{20.0\%} accuracy and improving OckScore to \textbf{17.06}. This shows that interpolation can prune redundant reasoning without retraining.

For the \textit{training-based} setting, we apply DIET \citep{chen2025overthinker}, an RL framework that adds difficulty-aware token penalties during fine-tuning. Starting from \texttt{DeepSeek-R1-Distill-Qwen-7B} and training on DeepScaleR, the optimized model improves accuracy from \textbf{14.5\%} to \textbf{15.5\%}, reduces average output length from \textbf{41,415} to \textbf{23,671} tokens, and raises OckScore from \textbf{7.39} to \textbf{10.23}. Table~\ref{tab:improvement} summarizes both results. Together, they show that reasoning efficiency is tractable to optimize through checkpoint merging or post-training alignment, and can be complemented by inference-time methods such as ThoughtTerminator \citep{pu2025thoughtterminator}.

\begin{table}[!h]
    \centering
    \caption{OckBench-Math efficiency improvements.}
    \label{tab:improvement}
    \small
    \setlength{\tabcolsep}{6pt}
    \renewcommand{\arraystretch}{1.15}
    \begin{tabular}{lccc}
        \toprule
        \textbf{Model Strategy} & \textbf{Acc (\%)} & \textbf{Avg Tokens} & \textbf{OckScore} \\
        \midrule
        \multicolumn{4}{l}{\textit{Interpolation Method}} \\
        Qwen3-4B-Thinking & 22.0 & 27,238 & 16.29 \\
        Qwen3-4B-Instruct & 13.5 & 11,859 & 10.10 \\
        \textbf{Qwen3-Mix-6:4 (Ours)} & \textbf{20.0} & \textbf{12,092} & \textbf{17.06} \\
        \midrule
        \multicolumn{4}{l}{\textit{RL Optimization Method}} \\
        DeepSeek-Distill-Qwen-7B & 14.5 & 41,415 & 7.39 \\
        \textbf{DeepSeek-DIET (Ours)} & \textbf{15.5} & \textbf{23,671} & \textbf{10.23} \\
        \bottomrule
    \end{tabular}
\end{table}

\section{Conclusion}

In this work, we introduced OckBench, a model- and hardware-agnostic benchmark for measuring \textit{Per-Token Intelligence}, the trade-off between reasoning accuracy and token consumption. Through the Differentiation Filter and the OckScore ($S_{Ock}$), OckBench moves beyond standard accuracy-only evaluation and directly measures how efficiently a model solves reasoning tasks.

We argue that the community should shift from accuracy-centric evaluation to per-token intelligence metrics. Models should be ``Ockhamized'' to organize reasoning pathways and stop treating tokens as unlimited. Our experiments on frontier and open-source models reveal token efficiency as a significant, overlooked axis of differentiation. Appendix~\ref{app:limitations_impact} discusses limitations and future work.

\bibliography{main}
\bibliographystyle{main}

\appendix
\clearpage
\section{Additional Benchmark Details}
\label{app:benchmark_details}

This appendix provides supplementary details for the construction and evaluation of OckBench. Its purpose is to make the benchmark design in Section~\ref{sec:benchmark_design} easier to reproduce and to clarify how OckScore should be interpreted in practice.

\subsection{Differentiation Filter in Practice}

The Differentiation Filter is designed to identify items that are informative for \emph{relative efficiency}, not merely raw capability. In practice, we first evaluate a diverse seed set of representative models on a large candidate pool. For each item, we record both correctness and output-token count. We then apply two stages of filtering:

\begin{enumerate}
    \item \textbf{Difficulty screening.} We remove items that are solved by nearly all models or by almost none of them, retaining only those whose average accuracy lies in a moderate band.
    \item \textbf{Variance ranking.} Among the remaining items, we rank candidates by the variance of output-token usage across models and keep the top-$k$ examples.
\end{enumerate}

This procedure intentionally favors questions that admit multiple viable reasoning paths. Such items are the most diagnostic for \textit{Per-Token Intelligence}: strong models tend to reach the solution with concise, targeted reasoning, whereas weaker or poorly aligned models often produce diffuse, repetitive chains before arriving at the same answer or timing out near a generation cap. The reported experiments use a frozen OckBench v1.0 split. Later applications of the same filtering recipe can produce refreshed benchmark versions, while v1.0 remains the longitudinal comparison point for the results in this paper.

\subsection{Interpreting OckScore}
\label{subsec:interpreting_ockscore}

The design of the unified \textit{OckScore} ($S_{\text{ock}}$) is fundamentally driven by a strict preference ordering for model deployment. We formalize our design philosophy as follows:
\begin{equation*}
    \langle \textsc{Correct}, \textsc{Short}\rangle \succ \langle \textsc{Correct}, \textsc{Long}\rangle \succ \langle \textsc{Wrong}, \textsc{Short}\rangle \succ \langle \textsc{Wrong}, \textsc{Long}\rangle
\end{equation*}

To capture this ordinal relationship quantitatively, OckScore is intended to preserve correctness as the primary objective while still penalizing unnecessary verbosity:
\begin{equation}
    S_{\text{ock}} = \text{Accuracy} - \lambda \cdot \log\left(\frac{T}{C} + 1\right)
\end{equation}
Here, $\text{Accuracy}$ is measured in percentage points (0--100), $T$ is the average number of output tokens, $\lambda$ is the token-penalty coefficient, and $C$ is a normalization constant. We use the natural logarithm throughout. In our experiments, we use $\lambda=10$ and $C=10{,}000$. For leaderboard aggregation, error cases are counted as incorrect, while average token usage is computed over non-error completions so that API or evaluator failures are not artificially penalized by missing token counts.

This formulation yields three practical consequences, which correspond directly to our core design considerations:

\paragraph{Accuracy Prioritization.}
Correctness is the primary utility of a reasoning model. By using accuracy as the base term and subtracting a logarithmic penalty, efficiency is treated as a \textit{cost}, not as a multiplier. This structure guarantees that a correct solution, no matter how verbose, generally outscores an incorrect one. Models cannot obtain a high score simply by being short; accuracy remains the dominant term, preserving the boundary between useful and useless models.

\paragraph{Logarithmic Scaling of Efficiency.}
We employ a logarithmic penalty on normalized token length ($\log(1+T/C)$) because reasoning lengths vary by orders of magnitude (e.g., 1k vs. 100k tokens). A linear penalty would either be negligible for short chains or disproportionately punitive for long, necessary reasoning. The logarithmic scale compresses this variance, discouraging token inflation without over-penalizing tasks that genuinely require deeper logical exploration.

\paragraph{Accuracy--Token Exchange Rate.}
Ignoring the zero-guard term, OckScore gives a fixed interpretation of token inflation: doubling the output length costs $\lambda\log 2$ accuracy points. Under our canonical $\lambda=10$, a model that uses twice as many tokens must gain approximately $10\log 2 \approx 6.9$ percentage points of accuracy to keep the same score. The exact implemented formula includes the $+1$ guard, so for a finite baseline token count $T$ the required gain is
\[
    \Delta \text{Accuracy}
    =
    \lambda \log\left(\frac{1+2T/C}{1+T/C}\right)
\]
which is slightly smaller for short traces and approaches $6.9$ points for long traces. For example, with $\lambda=10$ and $C=10{,}000$, doubling from 20k to 40k tokens costs 5.1 points, while the penalty for very long traces converges to the fixed $\lambda\log 2$ exchange rate.

\paragraph{Calibration with Model Priors.}
The parameters $\lambda=10$ and $C=10{,}000$ were calibrated to align with empirical priors regarding ``per-token intelligence.'' This parameterization ensures the score does not falsely promote ``short but dumb'' models. It yields a ranking where capable commercial models and larger parameter models within the same family score highest, reflecting their ability to condense complex reasoning into efficient paths. Conversely, smaller or weaker models are appropriately penalized for the ``overthinking tax'' described in Section \ref{sec:analysis}. Ultimately, this calibration makes comparisons between similarly accurate models highly informative: when two models achieve nearly the same Pass@1, the one with materially lower token consumption receives a meaningfully higher OckScore.

\paragraph{Sensitivity of $\lambda$.}
We treat $\lambda=10$ as an internal calibration, not as an optimal economic exchange rate between accuracy and tokens. On the 37-setting leaderboard in Table~\ref{tab:leaderboard_full}, two empirical signals bracket the choice. First, fitting the non-failure empirical Pareto frontier with
\[
    \text{Accuracy} = \alpha + \lambda \log(T/C+1)
\]
gives $\lambda \approx 20$ (95\% CI roughly 13--29), reflecting how much additional accuracy frontier models obtain as they spend more tokens. Second, 22 within-family ordering priors, such as preferring stronger generations or larger models within the same family, are best satisfied for $\lambda \approx 2$--6. The canonical value $\lambda=10$ sits between these signals. It satisfies 19 of 22 priors, only one fewer than the best attainable count of 20; two of the three violations are data-level dominance cases, not calibration failures.

\begin{table}[t]
    \centering
    \caption{\textbf{OckScore Sensitivity to $\lambda$.} Kendall $\tau_b$ is measured against the canonical $\lambda=10$ ranking on the 37-setting leaderboard. Rankings are stable around the canonical setting, especially for $\lambda \in [5,15]$.}
    \label{tab:ockscore_lambda_sensitivity}
    \small
    \setlength{\tabcolsep}{6pt}
    \begin{tabular}{lrrrrrr}
        \toprule
        $\lambda$ & 2 & 5 & 10 & 15 & 20 & 25 \\
        \midrule
        Kendall $\tau_b$ vs. $\lambda=10$ & 0.89 & 0.93 & 1.00 & 0.88 & 0.77 & 0.66 \\
        Priors satisfied & 20/22 & 20/22 & 19/22 & 19/22 & 16/22 & 17/22 \\
        \bottomrule
    \end{tabular}
\end{table}

This sensitivity analysis supports a limited claim: the main rankings and qualitative conclusions are not load-bearing on the exact value of $\lambda$ near 10. In particular, GPT-5.5 (medium) remains the top-ranked setting for all tested $\lambda \ge 10$. We do not claim that $\lambda=10$ is globally optimal, that it matches deployment pricing, or that the same calibration transfers unchanged to other benchmarks. Ignoring the zero-guard term, the reference constant $C$ contributes only a global additive offset in the conceptual form $\text{Accuracy}-\lambda\log(T/C)$; it fixes the score scale and does not serve as a second calibration parameter.

\subsection{Evaluation and Reproducibility Notes}

All evaluations use single-shot prompting and deterministic decoding wherever the serving API permits it. We count \textbf{output tokens} as the full generated response length prior to EOS, including both intermediate reasoning tokens and the final answer. This is the provider-observable generated-token budget reported by the API or counted with the native tokenizer for local models. Provider-private serving details are recorded when exposed by the endpoint, while the benchmark itself remains tied to the measurable token stream that users can audit and reproduce. This choice makes the metric directly reflect deployment cost, since latency and billing scale with generated tokens, not with the semantic usefulness of individual steps.

For answer checking, we use task-appropriate verification. Mathematical and science questions rely on final-answer extraction followed by exact matching against reference answers. Coding tasks are judged by execution against unit tests. This split is important: OckBench evaluates \emph{reasoning efficiency}, not stylistic preferences in intermediate traces. The benchmark therefore scores models according to whether they solve the task and how many generated tokens they require to do so.

The supplementary material provides the OckBench v1.0 selected-item files, runnable evaluation code, configuration templates, and thin per-setting result dumps for the 37-setting leaderboard. These artifacts omit full model response bodies while retaining correctness labels, extracted answers, token accounting, finish reasons, latency fields, and coding test outcomes, so the reported accuracy, token totals, and OckScore values can be recomputed from the released fields.

\subsection{Limitations and Broader Impact}
\label{app:limitations_impact}

\paragraph{Limitations.}
OckBench focuses on text-based math, coding, and science tasks where correctness can be checked reliably. This design keeps accuracy--token comparisons clean and directly interpretable. Extending the same methodology to multimodal reasoning, interactive agent workflows, and long-horizon software engineering is an important direction, since these settings introduce state, tool use, and partial progress that require richer evaluation protocols.

The Differentiation Filter is anchored by a representative set of models so that selected items lie near the current reasoning frontier. As model families advance, the benchmark should support dynamic refreshes: the same filtering procedure can be periodically rerun to keep OckBench focused on efficiency-revealing items.

Finally, model serving is a moving target. We use deterministic decoding where available, official token accounting when possible, native tokenizers for open-weight models, and the tokenizer sanity check in Appendix~\ref{sec:tokenizer_sanity_check}. Future benchmark releases can further improve longitudinal tracking by recording richer provider-snapshot and serving-stack metadata alongside the fixed prompt templates, raw accuracy, and token counts.

\paragraph{Broader impact.}
The main positive impact of OckBench is to make reasoning efficiency visible. Better token efficiency can reduce inference latency, user cost, and deployment energy use when models solve the same tasks with fewer generated tokens. This is especially relevant for long-running coding and agentic workflows where output length can dominate serving cost.

A possible concern for any efficiency metric is that it could be over-optimized if read in isolation. We therefore report raw accuracy and token counts alongside OckScore, keep correctness as the primary term, and treat OckBench as a complement to task-specific reliability and safety evaluations.

\subsection{Full Leaderboard}
\label{app:full_leaderboard}

For completeness, Table~\ref{tab:leaderboard_full} reports the full OckBench leaderboard, including multiple variants within the same model family that are omitted from the main-text summary table for space and readability.

\begin{table*}[t]
    \centering
    \setlength{\tabcolsep}{5pt}
    \renewcommand{\arraystretch}{1.2}
    \footnotesize
    \caption{\textbf{Full OckBench Leaderboard.} Models are ranked by OckScore ($S_{Ock}$). The table reports accuracy, average full-output tokens, and total output cost for the 200-item OckBench run (USD) from OpenRouter completion prices~\citep{openrouter_pricing_2026}. Shaded rows denote open-weight models.}
    \label{tab:leaderboard_full}
    \resizebox{\textwidth}{!}{
    \begin{tabular}{llrrrrrrrrrr}
    \toprule
    \multirow{2}{*}{\textbf{Model}} & \multirow{2}{*}{\textbf{Org}}
    & \multicolumn{2}{c}{\textbf{Coding}}
    & \multicolumn{2}{c}{\textbf{Science}}
    & \multicolumn{2}{c}{\textbf{Math}}
    & \multicolumn{2}{c}{\textbf{Overall}}
    & \multirow{2}{*}{\textbf{Cost}}
    & \multirow{2}{*}{\textbf{OckScore}} \\
    \cmidrule(lr){3-4} \cmidrule(lr){5-6} \cmidrule(lr){7-8} \cmidrule(lr){9-10}
    & & \textbf{Acc.} & \textbf{Tok.} & \textbf{Acc.} & \textbf{Tok.} & \textbf{Acc.} & \textbf{Tok.} & \textbf{Acc.} & \textbf{Tok.} & & \\
    \midrule
    GPT-5.5 (medium) & OpenAI & 96.7 & 1,739 & 85.0 & 4,043 & 80.0 & 6,723 & 86.0 & 4,692 & 28.2 & 82.2 \\
    GPT-5.5 (high) & OpenAI & 100.0 & 2,853 & 87.5 & 10,655 & 82.0 & 14,062 & 88.5 & 10,018 & 60.1 & 81.6 \\
    GPT-5.5 (xhigh) & OpenAI & 100.0 & 4,501 & 87.5 & 11,962 & 85.0 & 19,115 & 90.0 & 13,271 & 79.6 & 81.6 \\
    GPT-5.4 (medium) & OpenAI & 93.3 & 2,338 & 75.0 & 4,308 & 78.0 & 3,235 & 82.0 & 3,177 & 9.5 & 79.2 \\
    GPT-5.4 (high) & OpenAI & 98.3 & 3,877 & 70.0 & 7,698 & 74.0 & 659 & 80.5 & 3,267 & 9.8 & 77.7 \\
    Claude-Opus-4.7 & Anthropic & 85.0 & 1,878 & 85.0 & 6,076 & 81.0 & 11,406 & 83.0 & 7,481 & 37.4 & 77.4 \\
    GPT-5.4 (low) & OpenAI & 91.7 & 1,181 & 70.0 & 1,891 & 73.0 & 4,646 & 78.0 & 3,056 & 9.2 & 75.3 \\
    GPT-5.5 (low) & OpenAI & 95.0 & 738 & 70.0 & 1,272 & 65.0 & 2,255 & 75.0 & 1,603 & 9.6 & 73.5 \\
    Gemini-3.1-Pro & Gemini & 86.7 & 2,723 & 70.0 & 1,934 & 72.0 & 4,130 & 76.0 & 3,263 & 7.8 & 73.2 \\
    Claude-Opus-4.6 & Anthropic & 88.3 & 7,187 & 70.0 & 27,217 & 88.0 & 41,964 & 84.5 & 28,582 & 142.9 & 71.0 \\
    \rowcolor{black!10} DeepSeek-v4-flash (max) & DeepSeek & 96.7 & 32,709 & 75.0 & 90,834 & 77.0 & 111,211 & 82.5 & 83,585 & 4.7 & 60.1 \\
    \rowcolor{black!10} DeepSeek-v4-pro (max) & DeepSeek & 85.0 & 46,408 & 80.0 & 73,823 & 77.0 & 87,879 & 80.0 & 72,626 & 12.6 & 58.9 \\
    \rowcolor{black!10} Kimi-K2.6 & Kimi & 68.3 & 24,408 & 75.0 & 41,898 & 79.0 & 53,082 & 75.0 & 42,243 & 29.6 & 58.5 \\
    \rowcolor{black!10} Kimi-K2.5 & Kimi & 66.7 & 17,534 & 67.5 & 29,927 & 78.0 & 46,528 & 72.5 & 34,510 & 13.8 & 57.6 \\
    \rowcolor{black!10} Qwen3.5-397B-A17B & Qwen & 85.0 & 19,404 & 67.5 & 12,355 & 57.0 & 35,771 & 67.5 & 26,178 & 12.3 & 54.6 \\
    \rowcolor{black!10} MiniMax-M2.7 & MiniMax & 85.0 & 26,207 & 70.0 & 28,831 & 59.0 & 60,806 & 69.0 & 44,031 & 10.6 & 52.1 \\
    \rowcolor{black!10} MiMo-v2.5-Pro & MiMo & 75.0 & 14,569 & 60.0 & 30,384 & 61.0 & 48,868 & 65.0 & 34,882 & 20.9 & 50.0 \\
    Claude-Sonnet-4.6 & Anthropic & 88.3 & 12,471 & 45.0 & 27,366 & 46.0 & 13,973 & 58.5 & 16,201 & 48.6 & 48.9 \\
    \rowcolor{black!10} GLM-5.1 & GLM & 70.0 & 58,157 & 67.5 & 40,290 & 58.0 & 68,272 & 63.5 & 59,641 & 41.7 & 44.1 \\
    Gemini-3-Flash & Gemini & 55.0 & 2,088 & 57.5 & 2,343 & 33.0 & 1,076 & 44.5 & 1,615 & 1.0 & 43.0 \\
    Claude-Opus-4.5 & Anthropic & 53.3 & 1,092 & 60.0 & 837 & 32.0 & 1,146 & 44.0 & 1,068 & 5.3 & 43.0 \\
    \rowcolor{black!10} GLM-5 & GLM & 66.7 & 22,470 & 42.5 & 32,263 & 62.0 & 64,033 & 59.5 & 45,210 & 17.4 & 42.4 \\
    Gemini-2.5-Pro & Gemini & 53.3 & 10,964 & 70.0 & 14,463 & 47.0 & 28,468 & 53.5 & 20,979 & 42.0 & 42.2 \\
    \rowcolor{black!10} Qwen3.5-122B-A10B & Qwen & 68.3 & 13,874 & 55.0 & 14,045 & 44.0 & 33,166 & 53.5 & 23,554 & 9.8 & 41.4 \\
    \rowcolor{black!10} Gemma-4-31B-IT & Gemma & 51.7 & 4,374 & 45.0 & 3,118 & 34.0 & 3,681 & 41.5 & 3,776 & 0.3 & 38.3 \\
    \rowcolor{black!10} Qwen3.6-27B & Qwen & 46.7 & 51,678 & 55.0 & 22,443 & 54.0 & 70,673 & 52.0 & 55,329 & 35.4 & 33.2 \\
    \rowcolor{black!10} Qwen3.5-35B-A3B & Qwen & 41.7 & 13,769 & 55.0 & 13,878 & 45.0 & 72,916 & 46.0 & 43,364 & 8.7 & 29.3 \\
    \rowcolor{black!10} Qwen3.6-35B-A3B & Qwen & 40.0 & 47,195 & 52.5 & 52,959 & 50.0 & 65,809 & 47.5 & 57,655 & 11.5 & 28.4 \\
    Gemini-2.5-Flash & Gemini & 63.3 & 16,641 & 37.5 & 28,161 & 34.0 & 50,727 & 43.5 & 35,988 & 18.0 & 28.2 \\
    Claude-Sonnet-4.5 & Anthropic & 51.7 & 1,258 & 35.0 & 840 & 13.0 & 1,084 & 29.0 & 1,088 & 3.3 & 28.0 \\
    \rowcolor{black!10} Qwen3-235B-A23B & Qwen & 38.3 & 1,858 & 40.0 & 7,165 & 24.0 & 9,174 & 31.5 & 6,577 & 2.4 & 26.4 \\
    GPT-5.5 (none) & OpenAI & 38.3 & 141 & 35.0 & 9 & 15.0 & 431 & 26.0 & 260 & 1.6 & 25.7 \\
    \rowcolor{black!10} MiniMax-M2.5 & MiniMax & 41.7 & 66,984 & 62.5 & 16,824 & 39.0 & 67,849 & 44.5 & 57,346 & 13.2 & 25.4 \\
    Gemini-3.1-Flash-Lite & Gemini & 33.3 & 966 & 32.5 & 2,525 & 20.0 & 2,358 & 26.5 & 1,981 & 0.6 & 24.7 \\
    \rowcolor{black!10} MiMo-v2.5 & MiMo & 40.0 & 20,284 & 40.0 & 17,922 & 31.0 & 26,379 & 35.5 & 22,859 & 9.1 & 23.6 \\
    GPT-5.4 (none) & OpenAI & 20.0 & 122 & 45.0 & 241 & 17.0 & 1,209 & 23.5 & 689 & 2.1 & 22.8 \\
    \rowcolor{black!10} Qwen3.5-9B & Qwen & 23.3 & 48,894 & 50.0 & 37,287 & 9.0 & 200,282 & 21.5 & 116,222 & 3.5 & -3.9 \\
    \bottomrule
    \end{tabular}
    }
\end{table*}

\subsection{Tokenizer Sanity Check}
\label{sec:tokenizer_sanity_check}

A potential confounding factor when comparing reasoning efficiency is the inherent variation in tokenizers across different model families. To verify that our \textit{Output Tokens} metric is a reliable proxy for verbosity, we conducted a sanity check by applying the tokenizers of mainstream models to the complete text outputs (including long reasoning traces) of a single reference model evaluated on OckBench. The results indicate that the total token counts across these different tokenizers vary by at most 10\%. 

To ensure accurate counting in our main evaluation, we rigorously obtained token counts through the most direct and reliable paths for each model type:
\begin{itemize}
    \item \textbf{Proprietary Models:} We used the official token counting APIs or usage endpoints provided by the respective platforms (e.g., OpenAI, Google Vertex AI, Anthropic) to retrieve the exact token consumption.
    \item \textbf{Open-Weight Models:} We performed local counting using the native tokenizer configurations (e.g., official \texttt{tokenizer.json} or fast tokenizer implementations) from their public Hugging Face repositories, avoiding generic or unverified proxy tokenizers.
\end{itemize}

This minimal variance ($\le 10\%$) confirms that the massive differences in token consumption observed in our benchmark are driven by intrinsic model reasoning verbosity, not tokenizer vocabulary artifacts.

\subsection{Implementation Details for Efficiency Optimization}
\label{app:optimization_details}

In Section~\ref{sec:improvements}, we demonstrated that the token efficiency of reasoning models can be systematically improved. Here, we provide the detailed implementation configurations, hyperparameter selections, and mechanistic insights for the two optimization approaches evaluated in our study: training-free model interpolation and difficulty-aware reinforcement learning.

\subsubsection{Training-Free Optimization: Model Interpolation}
Model interpolation aims to combine the strong reasoning capabilities of ``Thinking'' models with the concise, instruction-following nature of ``Instruct'' models without the substantial computational overhead of fine-tuning. In our experiments, we used the \texttt{Qwen3-4B} family, specifically interpolating the weights of the \texttt{Qwen3-4B-Thinking} and \texttt{Qwen3-4B-Instruct} checkpoints.

We construct the merged model, \texttt{Qwen3-Mix}, using linear interpolation in the parameter space:
\begin{equation}
    \theta_{\text{merged}} = \lambda \cdot \theta_{\text{thinking}} + (1 - \lambda) \cdot \theta_{\text{instruct}}
\end{equation}

\textbf{Hyperparameter Selection and the Three-Stage Paradigm.} Following prior mechanistic insights on model interpolation~\citep{wu2025revisiting}, we did not select the mixing coefficient $\lambda$ arbitrarily. We recognize that interpolation follows a distinct three-stage evolutionary paradigm. A naive average ($\lambda = 0.5$) often falls into an unstable transition zone. Through systematic evaluation across $\lambda \in [0, 1.0]$, we identified that $\lambda = 0.6$ places the model at the onset of the stable reasoning phase (Stage 2/3 transition). This ratio retains the deep, exploratory logical structures of the Thinking model (e.g., feed-forward network traits) while strongly imposing the brevity constraints of the Instruct model. 

\textbf{Decoding Strategies.} Interpolated models can be sensitive to sampling parameters. To ensure rigorous evaluation, we conducted a grid search over temperature $T \in \{0.6, 0.65, 0.7\}$ and $\text{Top-}p \in \{0.8, 0.85, 0.9, 0.95\}$. We found that the merged model is remarkably robust, but adopting the Thinking model's native sampling parameters ($T=0.6$, $\text{Top-}p=0.95$) consistently yielded the most stable Pass@1 and Mean@1 metrics across our differentiation-filtered items.

\textbf{Results.} As shown in Table~\ref{tab:improvement}, this carefully calibrated interpolation effectively prunes redundant reasoning steps. The baseline \texttt{Qwen3-4B-Thinking} model averages 27,238 tokens per response with an accuracy of 22.0\% ($S_{\text{ock}} = 16.29$). By applying $\lambda=0.6$, \texttt{Qwen3-Mix-6:4} reduces the average output length by more than 55\% to \textbf{12,092} tokens. While accuracy has a minor dip to \textbf{20.0\%}, it remains higher than the pure Instruct baseline (13.5\%). The overall OckScore climbs to \textbf{17.06}, showing that models can be mathematically merged to achieve a better Pareto frontier of accuracy and token efficiency.

\subsubsection{Training-Based Optimization: Difficulty-Aware RL (DIET)}
For our training-based intervention, we leverage the DIET (Difficulty-Aware Training) framework \citep{chen2025overthinker}, which directly aligns the model to internalize reasoning efficiency via reinforcement learning (RL). We adopt \texttt{DeepSeek-R1-Distill-Qwen-7B} as our base model, as our main evaluation revealed it exhibits severe ``overthinking'' behavior, paying a high latency tax for poor accuracy.

\textbf{RL Objective and Advantage Weighting.} We conduct RL fine-tuning using the DeepScaleR dataset over the GRPO algorithm. A naive implementation of token penalties often collapses model performance because the penalty signal is distorted by the changing variance of outcome rewards across different difficulties. To ensure stable training, our implementation employs \textit{Advantage Weighting}. We normalize the advantages for the outcome reward and the token penalty independently before combining them:
\begin{equation}
    \hat{A}_{i}^{\prime} = \hat{A}_{\text{outcome},i} - \alpha_{\text{ada}}(x, \pi_{\theta}) \cdot \hat{A}_{p,i}
\end{equation}

\textbf{On-the-Fly Difficulty Estimation.} The coefficient $\alpha_{\text{ada}}$ is dynamically scaled based on the on-the-fly difficulty of the prompt, estimated by the pass rate of the samples within the current training batch. Queries classified as ``simple'' receive strict length penalties ($\alpha$ scales up to $\alpha_{\text{base}} = 0.5$) to discourage rambling, whereas ``complex'' queries are granted larger token budgets ($\alpha$ approaches $0$) to allow for necessary logical exploration.

\textbf{Results.} This targeted alignment yields substantial improvements across all evaluation dimensions. Post-training, the model's accuracy increases from \textbf{14.5\%} to \textbf{15.5\%}, indicating that constraining verbose hallucinations and encouraging concise logical pathways can benefit correctness. More importantly, the average token consumption is nearly halved, dropping from \textbf{41,415} to \textbf{23,671} tokens. The model's OckScore improves from \textbf{7.39} to \textbf{10.23}. These results validate that the ``Overthinking Tax'' is not an immutable characteristic of reasoning models, but a correctable behavior that can be systematically unlearned through properly structured, difficulty-aware RL incentives.

\end{document}